\documentclass[runningheads]{llncs}
\usepackage{graphicx}
\usepackage{subcaption} 
\usepackage{prettyref}
\newrefformat{eqn}{Equation~\ref{#1}}
\newrefformat{fig}{Figure~\ref{#1}}
\newrefformat{tab}{Table~\ref{#1}}
\newrefformat{sec}{Section~\ref{#1}}
\newrefformat{ssec}{Subsection~\ref{#1}}
\newrefformat{lst}{Listing~\ref{#1}}
\newrefformat{alg}{Algorithm~\ref{#1}}
\newrefformat{chapter}{Chapter~\ref{#1}}

\usepackage{float}
\usepackage{listings}

\newfloat{lstfloat}{htbp}{lop}

\begin{document}
\title{Dynamic Network selection for the Object Detection task: why it matters and what we (didn't) achieve. }
\titlerunning{Dynamic Network selection for the Object Detection task}
%
\author{Emanuele Vitali\inst{1} \and
Anton Lokhmotov\inst{2} \and
Gianluca Palermo\inst{1}}
\authorrunning{E. Vitali et al.}
%
\institute{Politecnico di Milano, Milano, Italy \\
\and
dividiti Ltd., Cambridge, UK\\
}
\maketitle              
\begin{abstract}
In this paper, we want to show the potential benefit of a dynamic auto-tuning approach for the inference process in the Deep Neural Network (DNN) context, tackling the object detection challenge.
We benchmarked different neural networks to find the optimal detector for the well-known COCO 17 database \cite{lin2014microsoft}, and we demonstrate that even if we only consider the quality of the prediction there is not a single optimal network. 
This is even more evident if we also consider the time to solution as a metric to evaluate, and then select, the most suitable network.
This opens to the possibility for an adaptive methodology to switch among different object detection networks according to run-time requirements (e.g. maximum quality subject to a time-to-solution constraint).

Moreover, we demonstrated by developing an ad hoc oracle, that an additional proactive methodology could provide even greater benefits, allowing us to select the best network among the available ones given some characteristics of the processed image.
To exploit this method, we need to identify some image features that can be used to steer the decision on the most promising network. 
Despite the optimization opportunity that has been identified, we were not able to identify a predictor function that validates this attempt neither adopting classical image features nor by using a DNN classifier. 

\keywords{Object Detection  \and Application Dynamic Autotuning.}
\end{abstract}

\section{Introduction}
A lot of progress has been done in the last 10 years in the context of Neural Networks. They have recently been used to solve complex problems such as image classification \cite{krizhevsky2017imagenet}, voice to text \cite{amodei2016deep} or object detection \cite{Girshick_2014_CVPR}. 
Since their introduction, they have eclipsed the old methods that were used to perform these tasks. 
In particular, they have become the de-facto standard in the field of computer vision for image classification and detection\cite{russakovsky2015imagenet}.

However, since there are a lot of different networks in literature, it is difficult to select the most suitable architecture (in terms of network deployed and hardware architecture used).
DNNs are characterized by an accuracy metric. In the object detection field, this metric is called mAP (mean average precision). This metric tells the user how accurate is the network in finding an object and classifying it.
This is not enough, since we may be interested in other characteristics of the network. Sometimes we have to run the network in resource-constrained devices, or we have to perform real-time classification, where the response time is important.
As an example in autonomous driving, an approximate detection in a short time is better than an accurate one that comes too late. 

An interesting job in classifying several networks by their accuracy and time to solution has been done in \cite{huang2017speed}. 
In this work, the authors classify some of the most important object detection networks and provide and compare their performances on a single GPU architecture.

Starting from that work, we benchmarked different networks on different CPU-GPU environments. From that experiment, we found out that there is no single one-fits-all network, even in terms of accuracy on a single image.
For this reason, we decided to analyze the problem of autotuning in this field, searching for some characteristics of the application or of the network itself that may enable a runtime network selection, whenever is beneficial. 
However, we were unable to find a suitable prediction function that can be used to drive the runtime selection, and thus to benefit from this optimization possibility.

The contributions of this paper can be summarized as follow:
\begin{itemize}
\item We performed a benchmarking campaign on different object detection networks aimed at exploring accuracy-performance tradeoffs; 
\item We demonstrate through a simple automotive use case how the dynamic autotuning approach can satisfy changing constraints that a single network was unable to satisfy;
\item We built an oracle function based on the benchmarking campaign, that can select the best network among the used ones for every image of the COCO17 dataset, thus evaluating the possible advantage in having a proactive (per image) autotuning approach;
\item We highlighted the failed attempts done to employ the proactive method either by finding some image features and training a selector using common machine learning techniques, or adopting an image classification network.
\end{itemize}

\section{Related Works}

Thanks to recent advances in the deep learning field, a lot of different models have been proposed to tackle the object detection challenge. These networks have different accuracy and execution times, and selecting the most suitable one is a complex task. 
Interesting approaches have been proposed considering the dynamic selection of networks for the context of image classification \cite{taylor2018adaptive,7331375,tann2018flexible,8741723}. 
In \cite{taylor2018adaptive}, the authors propose to select dynamically the image classification network performing the inference, proving that is possible to improve both the accuracy and the inference time thanks to an autotuning approach. There the authors use a K-Nearest-Neighbor predictor to select, among 4 different models, which one is the best to use for every different image. 
The usage of two networks with a big/LITTLE approach is proposed in \cite{7331375}. In this work, two different network architectures are created on a chip. One small and fast (the LITTLE architecture) and one that is more accurate and more time-consuming (the big architecture). They perform the inference with the little network and they use the big as a fallback solution only if the little network prediction is deemed not accurate. However, even in this work, the solution is proposed for the image classification challenge.
Another dynamic methodology for the image classification has been proposed in \cite{tann2018flexible}. Here the same network is trained several times, with different datasets, and an ensemble of networks is used to perform the inference. The networks are used sequentially and if a certain threshold metric is reached the result is returned without executing the remaining networks.
Several other design-time optimizations are proposed in literature to build the networks\cite{tan2019mnasnet}, to compress them \cite{10.1145/3210240.3210337} or to switch from image processing to more expensive and accurate input (eg. LIDAR) \cite{8741723}.
All of these work targeting the network selection are done in the context of image classification. Indeed, to the best of our knowledge, there is no work targeting the dynamic selection of the network in the object detection challenge.

\section{Motivations}

\begin{figure}[t]
	\centering
	\includegraphics[width=\columnwidth]{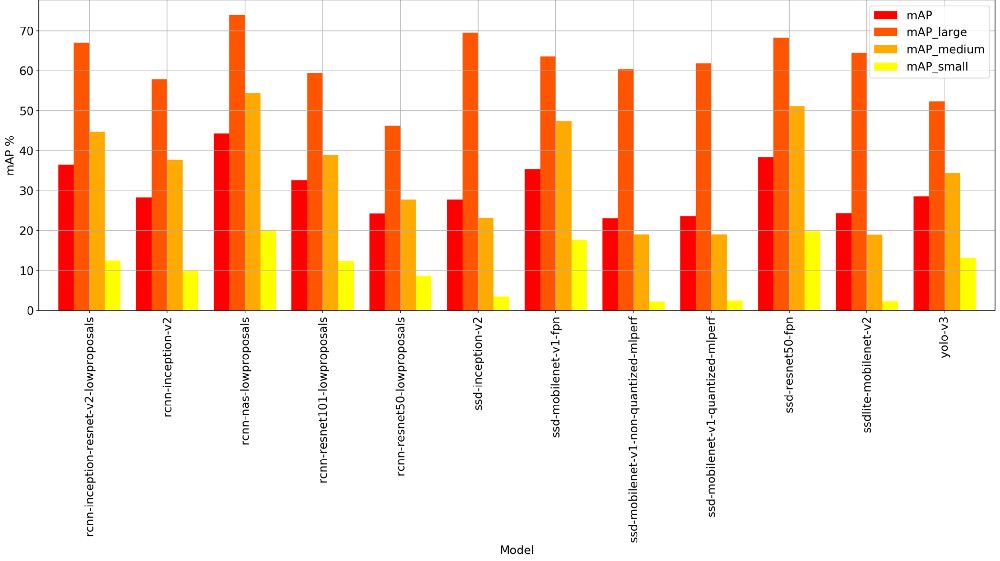}
	\caption{Results of the benchmarking accuracy campaign. }
	\label{fig:accuragy}
\end{figure}

To show the potential benefit of having a self-tuning network selector, we run an extensive benchmarking campaign on different object detection models and platforms. 
The objective of this campaign is to explore the behavior of different DNN on different platforms and with different configurations. In particular, we tested on CPU (with and without the AVX2 instructions) and GPU (with and without the TensorRT library support). 
We selected 12 different models. Most of them were coming from the Tensorflow Zoo \cite{tf_zoo}, trying to balance the SSD-based and the Faster-RCNN based models. To those models, we added a reimplementation of the YOLO-v3 network.

From the accuracy point of view, the campaign consists of 24 different experiments (12 models and with or without batch resizing). From the performance point of view, the number of experiments is increased to 360 and the whole Design of Experiment is reported in \prettyref{tab:config}.
The experiments have been done on the whole validation set of the COCO 2017 dataset.

\begin{table}[t]
\centering
\scriptsize
\begin{tabular}{|p{0.3\columnwidth}|p{0.6\columnwidth}|}
\hline
\textbf{Models} & Faster-rcnn-resnet50,
Faster-rcnn-resnet101, 
Faster-rcnn-NAS, 
Faster-RCNN-inception-resnetv2, 
ssd-mobilenet-v1-fpn, 
ssd-mobilenet-v1-quantized,
ssd-mobilenet-v1, 
ssd-resnet50-fpn, 
ssd-inception-v2, 
ssdlite-mobilenet-v2, 
rcnn-inception-v2, 
yolo-v3\\
\hline
\textbf{TF Configurations }& CPU, CPU with AVX2, GPU, GPU with TensorRT, GPU with TensorRT dynamic\\
\hline
\textbf{Batch sizes} & 1, 2, 4, 8, 16, 32 \\
\hline
\end{tabular}
\caption{Models used, configuration of Tensorflow and batching sizes used in the benchmarking campaign}
\label{tab:config}
\end{table}

As a motivation for the proposed idea, we will analyze the results of this benchmarking campaign, firstly from the accuracy point of view, then from the performance perspective and finally, we will analyze the Pareto frontier.

\begin{figure}[t]
	\centering
	\includegraphics[width=\columnwidth]{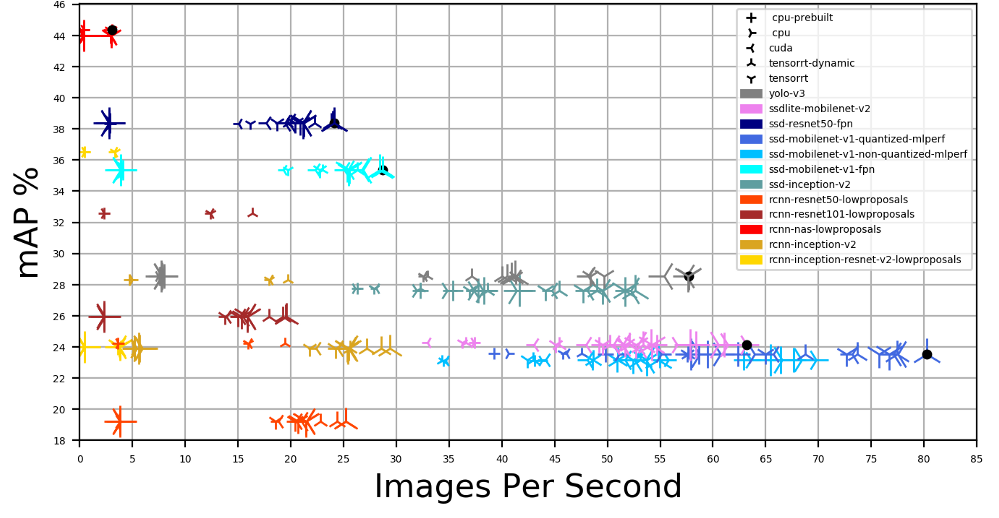}
	\caption{Result of the benchmarking campaign with the measured accuracy. The batch accuracy loss can be seen in the Faster-RCNN models. }
	\label{fig:full}
\end{figure}

\prettyref{fig:accuragy} shows the results of the accuracy benchmarking done while differentiating the accuracy also considering the \emph{size} of the object to be identified.
In the COCO dataset the objects are divided into three categories, small (up to 32*32 pixels), medium (from 32*32 to 96*96), and large (everything above).
The most accurate model is Faster-RCNN-NAS, which reaches the overall mAP of 44\%. Usually, a model with a good overall mAP performs consistently well across all three object categories. There are, however, some exceptions: SSD-Inception-v2 has the 2nd best score on large objects, but performs rather poorly on medium and small objects; on the contrary, YOLO-v3 has the 2nd worst score on large objects, but is on the 4th place on small objects and performs OK on medium objects.
The bad accuracy obtained on small objects is a well-known problem of SSD-based models. This is why the Feature Pyramid Network (FPN) feature have been introduced. Thanks to this feature, the SSD-ResNet50 and SSD-MobileNet-v1 models are able to reach 2nd and 3rd place on small objects (and on the 2nd and 4th place overall).

The complete result from the exploration can be seen in \prettyref{fig:full}. Here the color symbolizes the network used for the inference, while the shape is the backend and the size of the marker symbolizes the batch size. We can notice that the GPU backends are faster than the CPU ones and that the bigger points usually have the best performances. 

The images in the COCO dataset have different shapes and sizes. For the inference to be performed, they need to be resized to match the model input size.
This is usually done inside the network, as a first layer. However, this is not possible when processing a batch of images. In this case, all the images of a batch need to be resized before the inference is performed.
This procedure may damage the accuracy: some Faster-RCNN networks have the smallest point (no batch) at a higher accuracy level compared to the largest points of the same network.
Besides the Faster-RCNN-NAS, the other Faster-RCNN networks have a \textit{keep-aspect-ratio} layer, which becomes problematic when resizing the images to a unique size.
However, batching images can be significant for the performances, so we need to consider this possibility and not just discard it a-priori.
Indeed, as we can see from \prettyref{fig:full}, usually the bigger points have a better performance than the smallest one when we use the same backend. This growth can also be very significant, leading to almost double performances for some networks (YOLO v3 goes from 32 to almost 60 FPS).
However, the general behavior is not always true. Some networks show some unexpected results demonstrating how a dynamic selection of the most suitable configuration could be very important in this field.
The first is that batching can be detrimental to the performances: this happens when working with the Faster-RCNN-NAS on the CPU.
Another interesting result is that some networks perform better on the CPU than the GPU: an example is the ssdlite-mobilenet-v2 network

\begin{figure}[t]
	\centering
	\includegraphics[width=1\columnwidth]{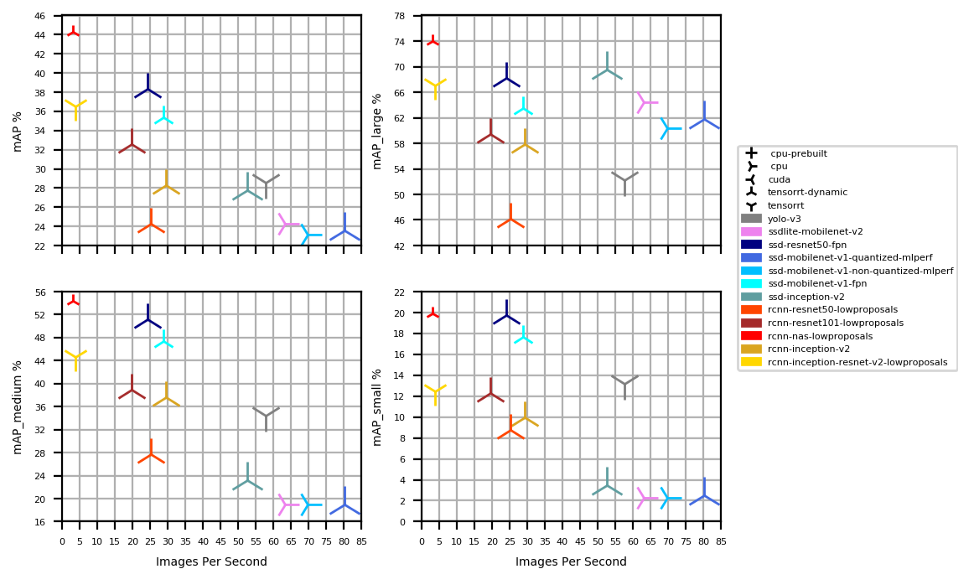}
	\caption{Best performance for every network at the different accuracy metrics (small, medium and large). }
	\label{fig:best}
\end{figure}

To conclude the motivational discussion, \prettyref{fig:best} shows the best configuration (considering ideal accuracy for the Faster-RCNN networks that have problems with batching). 
We can notice that there is not a one-fit-all optimal solution, since both the optimal backend and the optimal batch size changes across the different models.
Moreover, the networks on the Pareto set are also different if we consider different target accuracy.
All these variations strongly suggest that should a network selector function be found like in the methodology proposed in \cite{taylor2018adaptive} for the image classification challenge, the object detection challenge could largely benefit from an adaptive autotuning approach.

\section{The Proposed Approach}

\label{sec:proposedsolution}

In this section, we will see the methodology followed while trying to dynamically select the optimal inference network.
At the first time, we will see how exploiting two networks in analyzing a stream of frames can allow adapting to different constraints (maximize the accuracy of the prediction or maximize the frame rate) in a reactive way. 
With reactive, we mean that the autotuner reacts to the change of the constraints and responds to this change by selecting a different inference network, that can respect the new constraints. 
This approach follows the traditional reactive autotuning approach, used for example in \cite{filieri2015automated,socrate,baek2010green,mARGOt}.

\begin{figure}[t]
	\centering
	\begin{subfigure}{0.65\columnwidth}
		\includegraphics[width=\columnwidth]{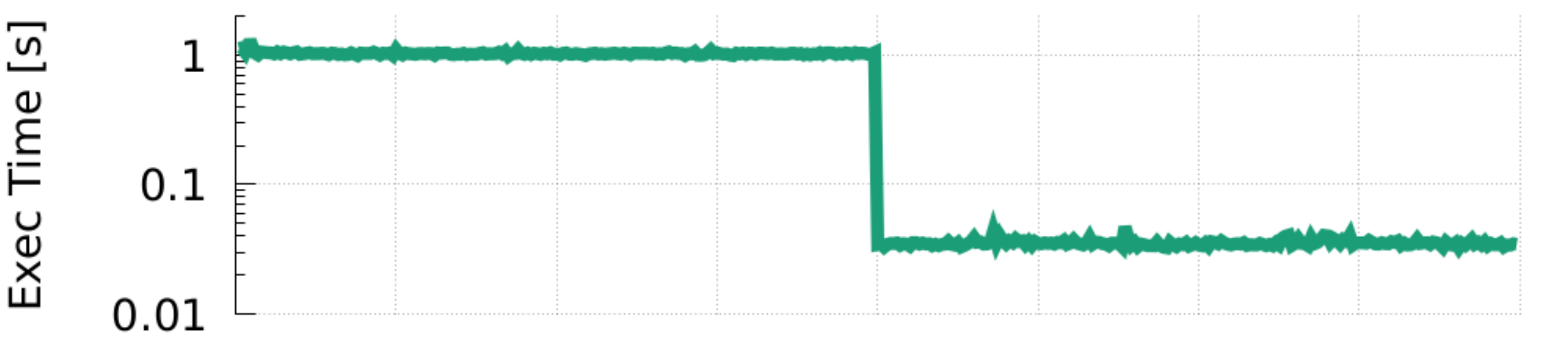}
		\caption{Execution time for every frame.}
		\label{fig:extime}
	\end{subfigure}
	\begin{subfigure}{0.65\columnwidth}
		\includegraphics[width=\columnwidth]{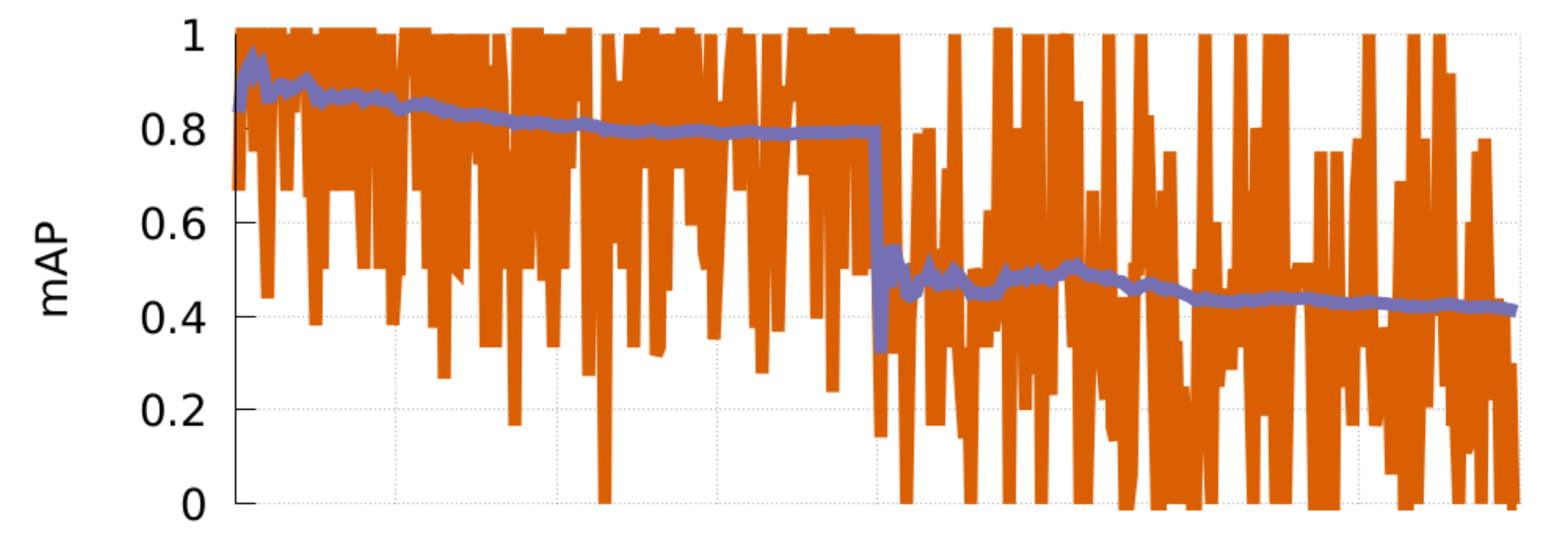}
		\caption{mAP of all categories}
		\label{fig:mapall}
	\end{subfigure}
	\begin{subfigure}{0.65\columnwidth}
		\includegraphics[width=\columnwidth]{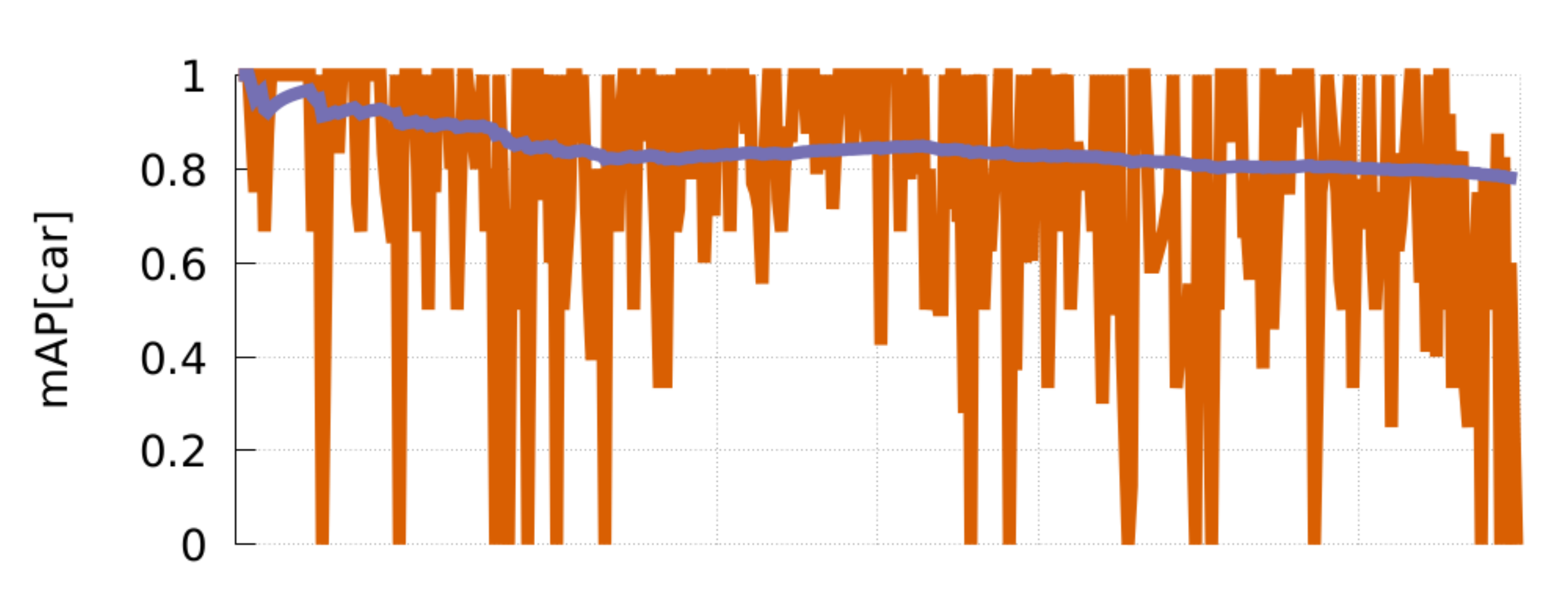}
		\caption{mAP of the car category}
		\label{fig:mapcar}
	\end{subfigure}
	\caption{Execution log of the same stream with two different networks, with a change of network in the middle. }
	\label{fig:traces}
\end{figure}

We then try to create a predictor function that can work as an oracle for unknown images. This is a proactive approach that relies on the concept of input feature that is present in some works in literature\cite{vitali2019efficient,laurenzano2016input}.
To create the predictor, we will search for some data features and we use them to create a function that can predict which is the best network to use to perform the inference.

\subsection{Reactive approach}
\begin{figure}[t]
	\centering
	\includegraphics[width=0.6\columnwidth]{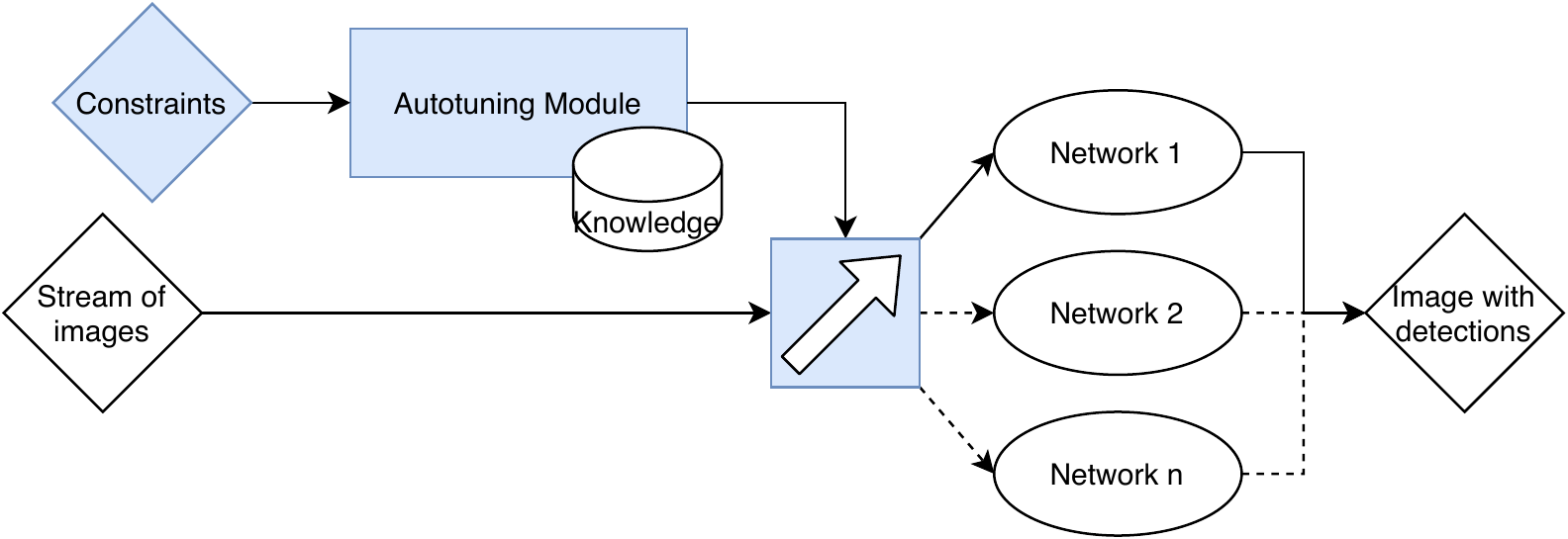}
	\caption{Architecture of the Reactive module. }
	\label{fig:reactive}
\end{figure}

In the reactive approach, the idea is of having the self-tuning module able to react to changes in the system or external conditions. Changes in external conditions are reflected in changes in the constraints.
\prettyref{fig:reactive} shows the approach from a high-level point of view: we must process a stream of images while respecting some constraints that may change during the runtime.
We have a set of networks with different (and known) characteristics in terms of accuracy and time to solution.
The autotuning module is in charge of selecting the most suitable among them according to the current constraints, interacting with factors that are external from the object detection problem.

To show the validity of this approach, let us suppose a very simplified scenario in the context of autonomous driving, which is one of the most important contexts in the object detection challenge.
We need to find the possible obstacles on or near the road, and we need to satisfy a strong constraint on the response time since if the detection arrives too late it is useless. 
To simplify the approach, let us suppose that we have 2 possible scenarios: highway and city driving. 
In the first case, we need to have a quick response and we need to identify "big" objects such as cars, while in the second case we have a slower speed, which means that we can use a slower network but we require a greater accuracy since we need to identify the "small" pedestrians. 

In this simplified example, the autotuner is in charge of switching from context 1 (city driving) to context 2 (highway) and back whenever a threshold speed is passed.
For this experiment, we have used the KITTI dataset \cite{Geiger2012CVPR}, which is a dataset created for the autonomous driving context.
As the first network (the fast one for the highway context), we retrained the SSDLite-Mobilenet, while as the accurate network we retrained the Faster-RCNN NAS network.

We show in \prettyref{fig:traces} an example of a run where we hypothesize to start the trip inside the city, where the most accurate network is used, and after a certain number of processed frames we move to a highway context, where we need faster processing.
We can notice from \prettyref{fig:mapall} that the mAP of the second network is noticeably worse than the first network.
However, we can also notice that the processing time of a single frame is almost two orders of magnitude faster.
If we look at \prettyref{fig:mapcar} instead, we can notice that the accuracy loss of the second network is slightly noticeable. In this image, we only considered the mAP obtained by the network in recognizing the car objects. 
In this way, we show that we can maintain the ability to find cars on the road within a constrained time to solution, which is smaller because of the higher navigation speed.
This result confirms the benefit of dynamic network selection in the context of the simplified scenario hypothesized before. Indeed, we can meet the accuracy/response time request in both the contexts, while both the considered networks are not able to do it if taken individually.

The impact of the reactive approach has been demonstrated on a simple use case considering only two networks. However, it can be easily generalized to a more complex one where, as an example, the constraints could be a function of the speed or multiple scenarios (and not only city and highway) can bring to different optimization problems.

\subsection{Proactive approach}
An orthogonal approach to the previous approach is the proactive one.
The proactive approach to dynamically select the network aims at using characteristics of both the network and the image that is going to be processed to match the image with its best possible network.
We believe that if there is not a one-fit-all best network while considering only the accuracy of the prediction, and there may be some features of the images that determine if a network behaves better than other networks in finding objects in that precise image.
Thus, we are interested in finding those characteristics of the images, and building a predictor that may be able to select the optimal object detection network.

\begin{figure}[t]
	\centering
	\includegraphics[width=0.60\columnwidth]{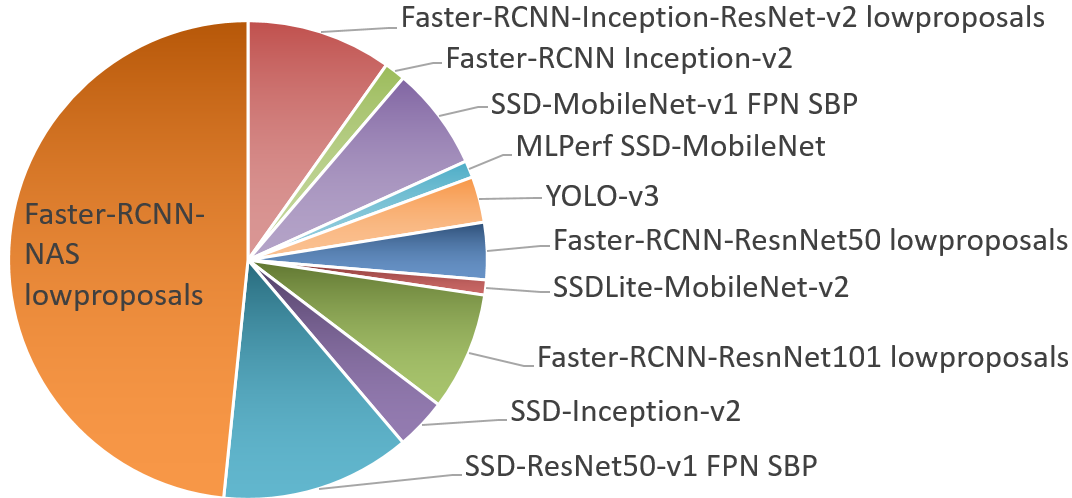}
	\caption{Composition of the Oracle on the full COCO validation set. All the considered networks are present, which means that they are optimal in predicting some images. }
	\label{fig:oracle}
\end{figure}

\begin{figure}[t]
	\centering
	\begin{subfigure}{0.6\columnwidth}
		\includegraphics[width=\columnwidth]{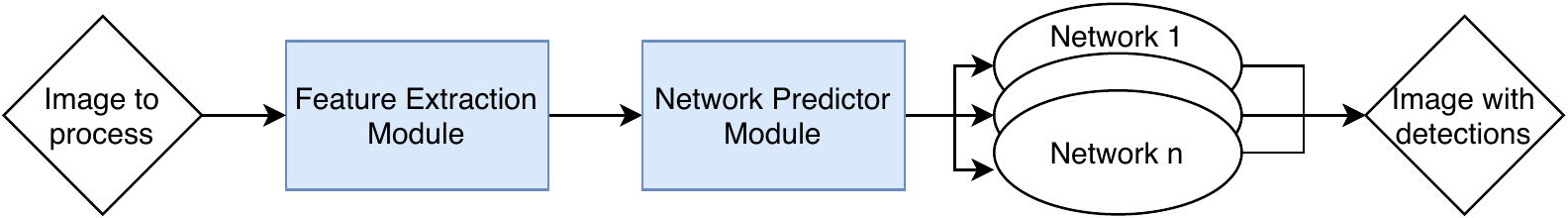}
		\caption{}
		\label{fig:predictor}
	\end{subfigure}
	\begin{subfigure}{0.6\columnwidth}
		\includegraphics[width=\columnwidth]{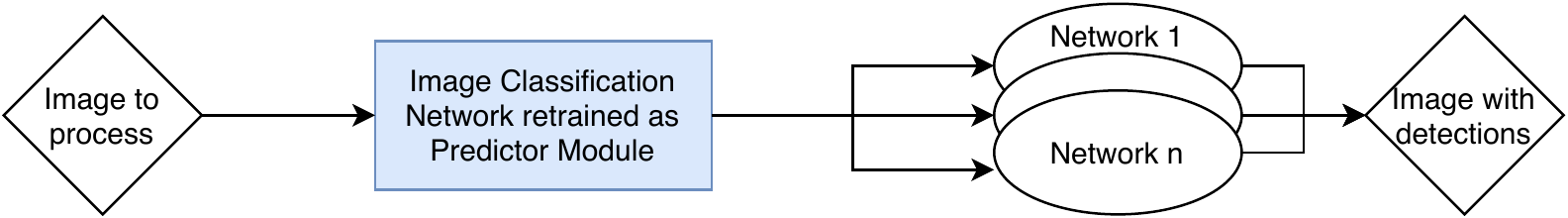}
		\caption{}
		\label{fig:classifier}
	\end{subfigure}
	\caption{Structure of the two attempts done in implementing the proactive approach to object detection, using a traditional Machine Learning approach (a) or using an Image Classification Neural Network (b).}
	\label{fig:attempts}
	\vspace{-5pt}
\end{figure}

The first step is to verify that the best network to perform the inference would change across the dataset. We create an oracle function, that selects the best network for all the images of the COCO validation set.
In particular, the program selects the highest mAP after evaluating an image with all the networks, and as a tiebreaker, it uses the execution time (the fastest one among the network with the same accuracy wins).
\prettyref{fig:oracle} reports the pie chart of the oracle. We can notice that almost half of the chart is occupied by the Faster-RCNN NAS, the most accurate network. We expected this network to be dominating. 
However, this network does not have always the best accuracy.
Moreover, the oracle shows that all the different networks are represented which means that they are optimal for at least some images.
The second step is to search the data features and a prediction function to drive the network selection proactively given the target image.
\prettyref{fig:attempts} shows two different attempts that we performed in building the predictor. The first one, which we define "traditional Machine Learning (ML) approach", can be seen in \prettyref{fig:predictor}. The second attempt, where we used neural network techniques, can be seen in \prettyref{fig:classifier}.
\prettyref{fig:predictor} shows the pipeline that we designed to perform object detection with network selection done with the traditional ML approach. The first step is Feature Extraction, which is a module that is in charge of quickly analyzing the image and extract some features. Then the predictor module is a function in charge of driving the network selection. This function needs to be able to quickly select the network given the data features extracted from the previous step.
Finally, the image is forwarded to the object detection network, which performs the detection task and returns the objects detected in the given image.
To create the feature extraction module, we need to identify a small set of features that can be quickly extracted from the image. 

We started the search of the data features from the ones used in \cite{taylor2018adaptive} since the authors were already working in the DNN context. 
Other candidate features are taken from \cite{laurenzano2016input}. In this work, four easily obtained characteristics  (\textit{mean, variance, local homogeneity, and covariance}) are used to decide how to approximate an image.
Moreover, we considered standard image processing features from literature \cite{hassaballah2016image}. 
We extracted all of these features and others using well-known Python packages, such as OpenCV \cite{2015opencv} and Skimage \cite{scikit-image}, collecting in total over 50 image features.
The complete list of the considered features is reported in \prettyref{tab:feats}.
\begin{table}[t!]
\scriptsize
\centering
\begin{tabular}{|p{0.32\columnwidth}|p{0.32\columnwidth}|p{0.32\columnwidth}|}
\hline
Number of keypoints & Number of corners & Number of contours\\
\hline
Dissimilarity & Homogeneity & ASM \\
\hline
Energy & Correlation & Number of peaks \\
\hline
Contrast & Variance & Mean\\
\hline
Hues(4 bins) & Saturation (4 bins)& Brightness (4 bins)\\
\hline
 Histogram of the three colors (3*8 bins) & Number of pixels that are edges in the image & Number of objects (connected components) \\
 \hline
Aspect ratio & Histogram of gradients(8 bins)&\\
\hline
\end{tabular}
\caption{List of all the features collected to build the predictor.}
\label{tab:feats}
\end{table}
We did extract all of these features, however, we are aware that we need to reduce the number of features to use, since getting all of these would be too time-consuming.
Moreover, some of them (for example the connected components) are too expensive in terms of extraction time and have been discarded a priori.

\begin{figure}[t]
	\centering
	\begin{subfigure}{0.65\columnwidth}
		\includegraphics[width=\columnwidth]{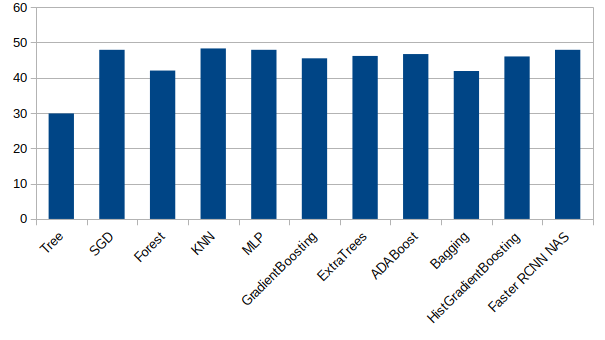}
		\caption{Accuracy of the predictors with all the networks}
		\label{fig:11nets}
	\end{subfigure}
	\vspace{-0.9pt}
	\begin{subfigure}{0.65\columnwidth}
		\includegraphics[width=\columnwidth]{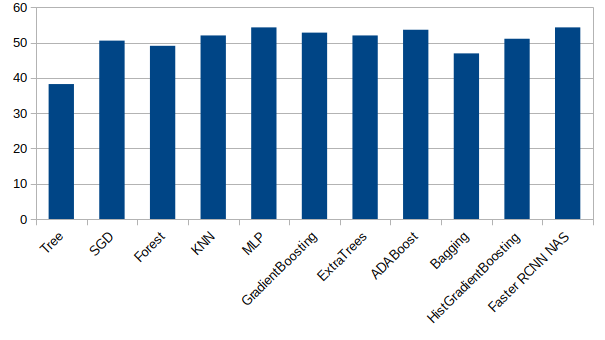}
		\caption{Accuracy of the predictors with a restricted set of networks.}
		\label{fig:6nets}
	\end{subfigure}
	\caption{Results of the training of the different models.}
	\label{fig:predictors_results}
\end{figure}

The following step is to build the classifier. To do this, we use both the output of the oracle and the extracted features of the images, since we need to learn the correlation between these features and the best network.
We decided to use the scikit library \cite{scikit-learn} since it is a well-known and verified module for the most common ML algorithms.
We used a Principal Component Analysis (PCA) to restrict the space of features, assigning to this methodology the duty of finding out which ones are the most important features that we have to consider. 
We then passed the output of the PCA to the following step, which is the model training. Before training the model, we have to create the training and the test set. 
From the available 5000 images (for which we have the array of features with the associated best network), we create a training set of 4500 images, while the other 500 are left as validation set.
Since the goal is to implement a classification layer, we have tested most of the classifier engines available in the \textit{scikit-learn} module. Among them, we tested \textit{Decision Tree, Random Forest, Bagging, AdaBoost, Extra Trees, Gradient Boosting, SGD, MLP, KNeighbors}.
However, no one of those algorithms was able to provide a robust classifier that could be used as the predictor, as we can notice from \prettyref{fig:predictors_results}. 
In particular, \prettyref{fig:11nets} shows the result on the complete set of networks. In most cases, the accuracy of the validation set was around 40\% which is also the number of occurrences of the most accurate model always (the last column in the figure). The tree predictor is the one that shows the worst result, around 30\%. 
To reduce the noise in the data available for the learning phase, we restricted the number of models. We decided to use only the ones that were Pareto optimal in the benchmarking study. This reduced the number of available models to 6.
Nonetheless, even with the reduced number of target networks, the traditional ML classifiers were unable to learn how to predict the best network to use to perform object detection given the image. The result of this final experiment is reported in \prettyref{fig:6nets}. We can notice that even with this reduction in the possible networks there is no valid predictor: the last column (Faster-RCNN-NAS) is the predictor that always selects the Faster-RCNN-NAS network to perform the detection since it is the most accurate one. This predictor has an accuracy of 55\%, which means that in more than half of the test images the RCNN-NAS has the optimal accuracy in the reduced validation set. 
All the predictors have a worse result, meaning that they can guess the optimal network with less accuracy than always selecting the same, and most used, network.

Since the traditional approach did not lead us to a working solution for our problem, we decided to attempt using a DNN classifier. In particular, we selected a MobilenetV2, trained on the ImageNet dataset. We decided to perform transfer learning, thus only modifying the last layers (the classifier layers) of the network, without changing the feature extraction layers.
The network we used to perform the transfer learning has 154 frozen layers and the last layer has 1280 features coming out, to which we attach the dense layers used to perform the classification.
The total number of parameters of this network is 5,149,771, and more than half of them are frozen, so they cannot be trained during the transfer learning.
As we can see, we have much more features than with the previous approach.
We used the keras \cite{chollet2015keras} framework to perform the transfer learning.
Since the oracle shows that there is no a similar amount of images for all the network, we needed to rebalance the dataset to have a fair training phase. The training data have been preprocessed to obtain a balanced dataset where all the labels (in our case, the target networks) have the same amount of training images.
This is a well-known technique used to avoid that
the dataset unbalancing can influence the learning process.
However, even this approach did not lead to a working predictor. The new predictor always learns to predict one or two networks.

We do not know the exact reason behind all of these failures.
We believe that the main reason is that object detection is a much more complex operation if compared to image classification 
where a similar attempt was successful \cite{taylor2018adaptive}. 
Indeed, the DNN used to tackle this challenge are more complex than the classification networks:
\cite{huang2017speed} shows how most object detection networks are composed of two sections, a region proposal network that aims at creating the bounding boxes of the objects, and a feature extractor, which is an image classification network that provides the label to the object extracted with the first stage.
We think that this failure may be due to the fact that the image features extracted with traditional image processing or with feature extraction layers trained for the classification problem are not enough. Indeed, these features may not be sufficient to model the region proposal problem. 
Thus, a different set of features may be needed. 

\vspace{-5pt}
\section{Conclusion}
In this paper, we studied the possibility of dynamically select the network used to perform inference in the object detection context, where to the best of our knowledge has not been already attempted before. 
We have shown why a dynamic autotuning methodology could be very profitable for this context, with a large benchmarking campaign that demonstrates that there is no a one-fit-all optimal solution.
We have seen that the use of a reactive approach can satisfy changing requirements by exploiting networks that were unable to satisfy the given constraints if taken singularly.
We tried also to adopt a proactive approach, that could have been even more profitable if we were able to select in advance the most suitable network for a specific image. While the oracle confirmed our feeling, we were not able to find any feature extraction technique that was able to drive the selection process.
Finally, we believe that this work is a good motivational study and our attempts could be useful by other researchers interested in the field.

\subsubsection*{Acknowledgements}
Part of this work was done when Emanuele was an intern in dividiti Ltd, Cambridge.
The internship was sponsored by the HIPEAC project, in the European Union’s Horizon 2020 research and innovation programme under grant agreement number 871174.
Moreover, part of this work has been developed within EVEREST project and funded by the EU Horizon 2020 Programme under grant agreement No 957269.
%
%
%
%
\bibliographystyle{splncs04}
\bibliography{biblio}
\end{document}